\documentclass[fleqn,10pt]{wlscirep}
\usepackage[utf8]{inputenc}
\usepackage[T1]{fontenc}
\usepackage{soul}
\usepackage{url}

\title{Learning by Active Forgetting for Neural Networks}

\author[1]{Jian Peng}
\author[2]{Xian Sun}
\author[1]{Min Deng}
\author[1]{Chao Tao}
\author[3]{Bo Tang}
\author[4]{Wenbo Li}
\author[1]{Guohua Wu}
\author[5]{Qing Zhu}
\author[6]{Yu Liu}
\author[7]{Tao Lin}
\author[1]{Haifeng Li*}

\affil[1]{School of Geosciences and Info-Physics, Central South University, Changsha 410086, China}
\affil[2]{Chinese Academy of Sciences, Aerospace Information Research Institute, department, Beijing 100094, China}
\affil[3]{Department of Electrical and Computer Engineering, Mississippi State University, Starkville, MS 39762 USA}
\affil[4]{Institute of Intelligent Machines, Chinese Academy of Sciences, Hefei 230031, China}
\affil[5]{Faculty of Geosciences and Environmental Engineering, Southwest Jiaotong University, Chengdu 611756, China}
\affil[6]{Institute of Remote Sensing and Geographic Information System, Peking University, Beijing 100094, China}
\affil[7]{College of Biosystems Engineering and Food Science, Zhejiang University, Hangzhou 310058, China}
\affil[3]{corresponding.author lihaifeng@csu.edu.cn}

\begin{abstract}
	
Remembering and forgetting mechanisms are two sides of the same coin in a human learning-memory system. Inspired by human brain memory mechanisms, modern machine learning systems have been working to endow machine with lifelong learning capability through better remembering while pushing the forgetting
as the antagonist to overcome. Nevertheless, this idea might only see the half picture. Up until very recently, increasing researchers argue that a brain is born to forget, i.e., forgetting is a natural and active process for abstract, rich, and flexible representations. This paper presents a learning model by active forgetting mechanism with artificial neural networks. The active forgetting mechanism (AFM) is introduced to a neural network via a "plug-and-play" forgetting layer (P\&PF), consisting of groups of inhibitory neurons with Internal Regulation Strategy (IRS) to adjust the extinction rate of themselves via lateral inhibition mechanism and External Regulation Strategy (ERS) to adjust the extinction rate of excitatory neurons via inhibition mechanism. Experimental studies have shown that the P\&PF offers surprising benefits: self-adaptive structure, strong generalization, long-term learning and memory, and robustness to data and parameter perturbation. This work sheds light on the importance of forgetting in the learning process and offers new perspectives to understand underlying mechanisms of neural networks.
	
\end{abstract}
\begin{document}

\flushbottom
\maketitle
%
%
\thispagestyle{empty}


\section*{Introduction}
Learning and memory capability are the essence of biological intelligence.
Questions about how the brain processes information, encodes, and stores knowledge in learning and memory have increasingly attracted considerable research \cite{craik1992human,piaget2015memory} and achieved remarkable progress in recent decades \cite{Hassabis2018Neuro},
relighting the hope toward artificial general intelligence in a brain-inspired way. In particular, inspired by the memory mechanisms of the human brain, modern machine learning systems have been working to endow machines with lifelong learning capabilities by overcoming catastrophic forgetting in neural networks \cite{kirkpatrick2017overcoming}, which intuitively furthers the long-held view of forgetting as a negative consequence of memory and as a passive process  \cite{loftus1985evaluating,carpenter2008effects}. Until recent research challenge this opinion: a brain is born to forget, i.e., forgetting is a natural and active process for abstract, rich, and flexible representations \cite{gravitz2019forgotten}. Forgetting also plays a crucial role in regulating the learning-memory process for good generalisability in the real-world \cite{bekinschtein2018retrieval}. \emph{"If we remember everything, we should on most occasions be as ill off as if we remember nothing."} This quote, cited by William James \cite{james2007principles}, concludes that forgetting is essential to humans to move forwards.

Deep neural networks (DNNs), a technique widely used for a range of machine intelligence-related tasks,  are essentially a form of forgetting that compresses knowledge by filtering irrelevant information from massive signals \cite{zoph2016neural,tishby2015deep}, and many of its techniques draw on the abstract concept of forgetting \cite{lecun2015deep}. Popular implicit approaches, like drop out \cite{krizhevsky2012imagenet}, L1 norm \cite{zhang2016l1}, and decorrelation \cite{cogswell2015reducing}, regularise neurons' activation, promoting the model’s strong generalization, robustness, or low parameter complexity \cite{yang2011robust,yang2018scalable}. Alternatively, explicit works specify individual modules, e.g., LSTM \cite{yang2011robust,yang2018scalable} and its variants \cite{yang2011robust,yang2018scalable}, and gated control \cite{liu2016spatio} networks to filter spatial-temporal contextual information. Furthermore, some works mimics mechanisms of natural forgetting, e.g., using the lateral inhibition mechanism of visual cells to suppress category-level irrelevant features in saliency detection \cite{cao2018lateral} or improve extended sequence memory capacity in lifelong learning \cite{aljundi2018selfless}. In general, these efforts suggest that forgetting potentially plays an essential role in learning and memory in artificial neural networks. Nevertheless, research on artificial neural network-based forgetting mechanisms and their explicit modelling is missing.

In contrast to artificial neural networks, forgetting mechanisms in biological neural networks have been extensively studied in recent years \cite{hardt2013decay,davis2017biology,tononi2014sleep}. Growing evidence \cite{shuai2010forgetting} suggests that the brain actively forgets through the selective extinction of neurons, which plays a fundamental role in the learning-memory process \cite{davis2017biology,izawa2019rem}. Further studies \cite{shuai2010forgetting} suggest that this regulation of forgetting is accomplished through a class of molecular switches, namely neurons or proteins, that selectively regulate the activation of excitatory neurons. Concretely, these switches are semi-open before learning and adjust to both directions, i.e., closed during learning, with a corresponding acceleration of forgetting, and open during memory consolidation, with a corresponding slowing of forgetting conversely. Furthermore, the rate of forgetting depends on the degree of switching, which is regulated according to a competitive mechanism that cooperates with inhibition and lateral inhibition to sparsely encode the gist, allowing the brain to learn and remember more events \cite{yu2014sparse}.

Here, we propose a memory model for artificial neural networks by active forgetting mechanism inspired by the human brain. We design a "plug-and-play" forgetting layer (P\&PF) consisting of groups of inhibitory neurons with Internal Regulation Strategy (IRS) to adjust the extinction rate of themselves by mimicking biological lateral inhibition mechanism and External Regulation Strategy (ERS) to adjust the extinction rate of excitatory neurons by mimicking biological inhibition mechanism.
We demonstrate that this neural switch-based biological forgetting mechanism is also applicable to artificial neural networks. It promotes strong generalization, long-term memory, and strong robustness for data/weight perturbations.

\begin{figure}[!htbp]
\centering
\includegraphics[width=0.8\linewidth]{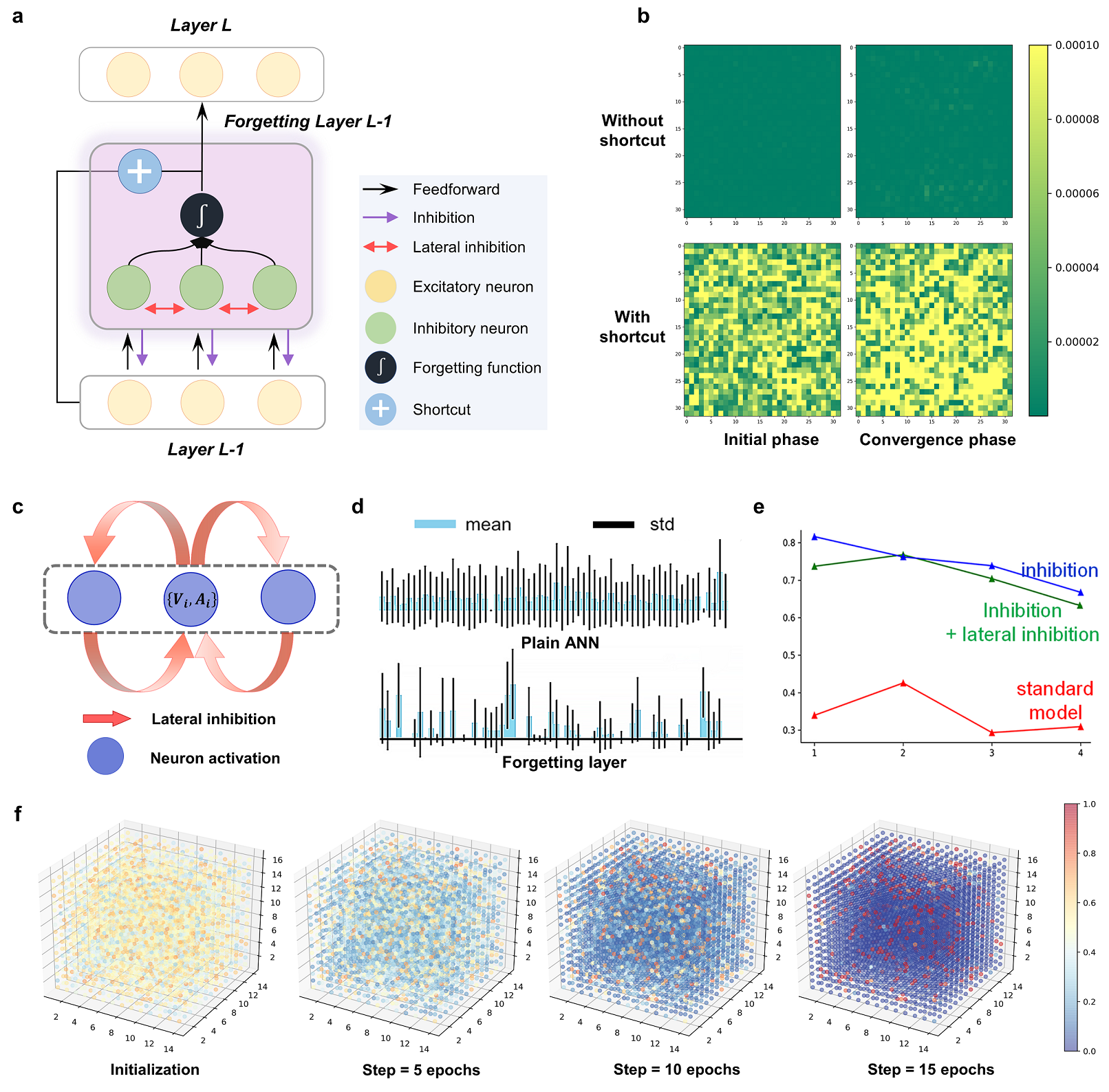}
\caption{Schematics of the proposed forgetting layer in artificial neural networks. \textbf{(a).} The structure of the forgetting layer in artificial neural networks, which lies between two adjacent layers of excitatory neurons. The signalling of excitatory neurons contains two paths: one outputs the response of excitatory neurons to the next layer of excitatory neurons after the forgetting operation of inhibitory neurons; the other directly connects to the next layer through a short cut. \textbf{(b).} Gradient values of excitatory neurons in the intermediate layers during the initial and convergence phases of training, with or without the shortcut. Colder colours indicate lower gradient values and vice versa. \textbf{(c).} The lateral inhibition mechanism of inhibitory neurons in the forgetting layer, based on their learned importance. \textbf{(d).} Activation of excitatory neurons for standard neural networks with and without forgetting layers. The horizontal axis represents neurons; the solid blue line and the solid black line are the mean activation value and standard deviation of the samples counted in the test dataset. \textbf{(e).} Comparison of the sparsity of activation values of excitatory neurons using different mechanisms on a four-layer multilayer perceptron, including plain artificial neural networks, ERS, and cooperative IRS and ERS. \textbf{(f).} The process of selective apoptosis of excitatory neurons during active forgetting. Warmer colours indicate more intense closure of the corresponding inhibitory neurons. The excitatory neuron is primarily active in the initialization phase. Then, the majority of excitatory neurons are gradually deactivated during training.}
\label{fig1}
\end{figure}

\section*{Results}

\subsection*{Actively Forgetting via the Plug-and-Play Forgetting Layer.}
Fig. \ref{fig1}a gives a schematic architecture of the active forgetting model on an artificial neural network, i.e., adding a plug-and-play forgetting layer (P\&PF), consisting of a group of inhibitory neurons regulated by ERS and IRS.
More specifically, ERS mimics the biological inhibition mechanism by compressing the size of inhibitory neurons through the softmax function with a soft threshold, which acts as a bottleneck that allows only strong inhibitory neurons to survive (see Methods for details). IRS mimics the biological lateral inhibition mechanism, which strengthens vital inhibitory neurons and weakens weak inhibitory neurons through mutual inhibition between inhibitory neurons (Fig. \ref{fig1}c), which is complementary to ERS.

We found that the forgetting layer is difficult to train and converge because of the oversaturation problem of the softmax function, which causes gradients to vanish, as shown in the top row of Fig. \ref{fig1}b. Drawing on identity \cite{he2016deep}, a shortcut is designed between the activation and forgetting layers to prevent gradient disappearance. The shortcut can effectively propagate the gradient on parameters compared with no shortcut, allowing more parameters to adjust in a broader range during the whole model training process.

We first compare the activation states of excitatory neurons on a multilayer perceptron (MLP, with 748-128-256-10 units) with and without the P\&PF in Fig. \ref{fig1}d. The results demonstrate that using the forgetting layer can substantially decrease excitatory neurons' activation and achieve a sparse representation. For quantification, referring to the metric of sparsity in biological neurons \cite{yu2014sparse}, we propose a metric to evaluate the sparsity in artificial neural networks. It is calculated by
\begin{equation}\label{eq:1}
    s =\left\{1-\frac{\left[\sum_{i=1}^{N}\left(\frac{r_{i}}{N}\right)\right]^{2}}{\sum_{i=1}^{N} \frac{r_{i}^{2}}{N}}\right\}\left(1-\frac{1}{N}\right)
\end{equation}

where $N$ denotes the number of units in one layer, and $r_{i}$ denotes the activation frequency of the $i$-th neuron on all test datasets.

We demonstrate that P\&PF substantially enhances the sparsity of excitatory neurons. Fig. \ref{fig1}e compares the sparsity of the standard MLP and MLP with P\&PF using the inhibition mechanism (ERS) and the lateral inhibition mechanism (IRS), showing that the sparsity of the model using the forgetting layer is much higher than the plain model. Furthermore, forgetting is selective in layers. The upper layer’s sparsity is considerably lower than the lower layer, and it does the same in standard models. It might be because the representation of higher layers is more complex than that of lower layers, which requires more neurons to encode. Introducing a lateral inhibition mechanism may reduce sparsity, but it vastly improves the generalized accuracy, implying a more efficient way of encoding representations. In addition, we visualize the intensity changes of neural switch-inhibitory neurons (Fig. \ref{fig1}f) to analyze the extinction process of excitatory neurons regulated by the P\&PF.
Prior to learning, the strength of inhibitory neurons is at an intermediate value, implying massive excitatory neurons join in encoding; then, inhibitory neurons increase rapidly during learning, which suppresses the expression of excitatory neurons insignificant to the task; finally, the majority of inhibitory neurons turn on at the convergence stage, which corresponds to the extinction of a vast number of activating neurons.

\begin{figure*}[t]
\centering
\includegraphics[width=.9\linewidth]{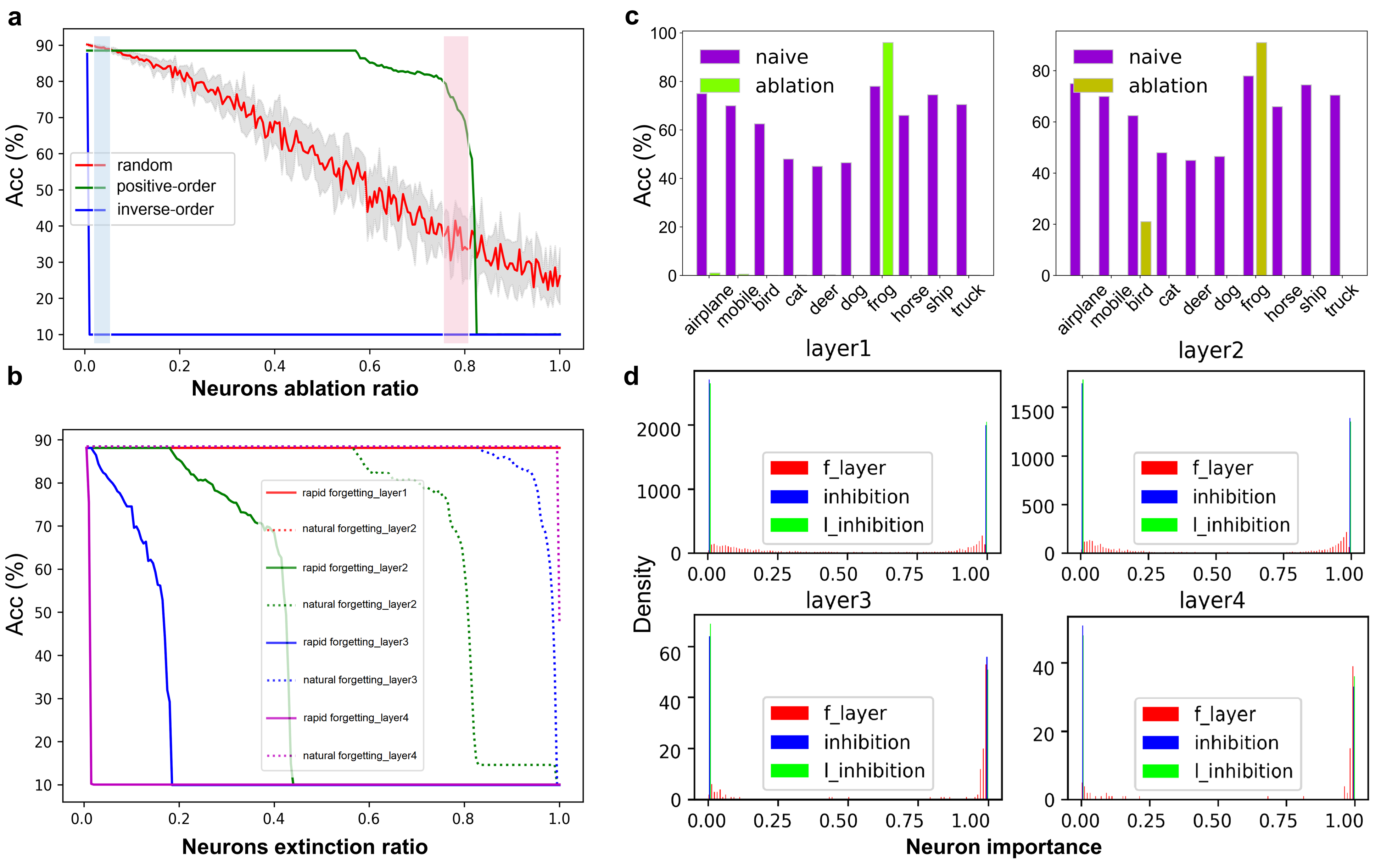}
\caption{The forgetting layer learns the neuron-importance of excitatory neurons. \textbf{(a).} Results of ablating excitatory neurons. The horizontal axis indicates the proportion of neuron ablations, and the vertical axis indicates the model test accuracy. Three types of ablation methods are utilized: random (red line is the mean of test accuracy for ten operations, and shading is the result considering standard deviation), positive-order ablation according to importance, and inverse-order ablation according to importance.  \textbf{(b),} Results of natural and rapid forgetting in different layers using the forgetting layer.
 \textbf{(c).} The model identifies the results on each class after removing neurons with top-1 and top-10 importance values. \textbf{(d).} The distribution of neuron-importance values of excitatory neurons in each layer of the network. Three methods are compared: forgetting layers with and without using a cooperative mechanism of inhibition and lateral inhibition, as well as forgetting layers using only an inhibition mechanism.}
\label{fig2}
\end{figure*}

\begin{figure*}[!htpb]
\centering
\includegraphics[width=.95\linewidth]{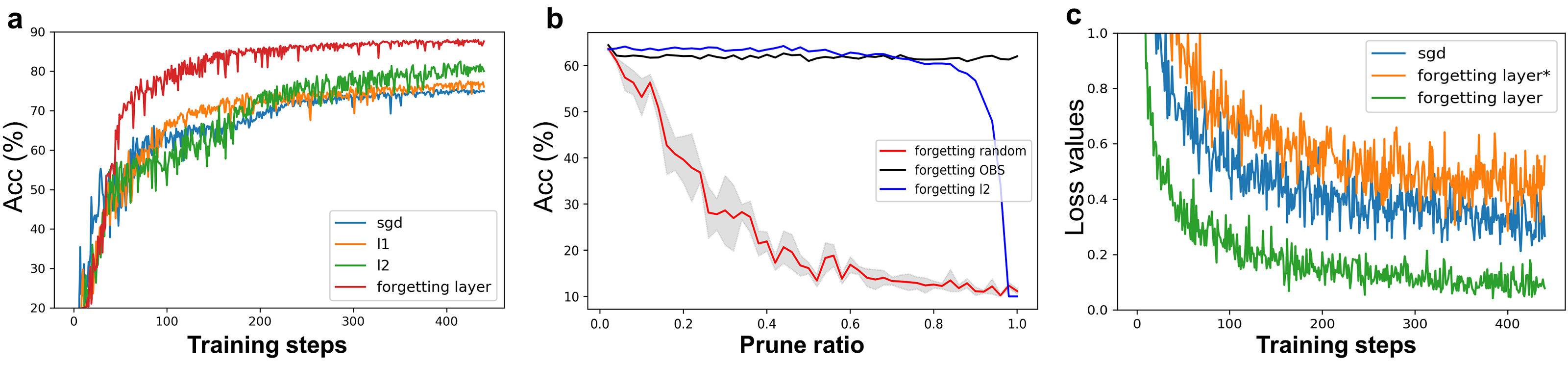}
\caption{Facilitating efficient learning in artificial neural networks using forgetting layers. \textbf{(a).} Structural adaptive information encoding. Separate models are trained on Fashion MNIST for the vanilla multilayer perceptron ($sgd$), the model with L1 regularization($l1$), the model with L2 regularization ($l2$), and the model with the forgetting layer ($forgetting layer$). \textbf{(b).} Comparison of the actual parameter occupancy of the models. Three parameter pruning methods are applied: random pruning ($forgetting random$), optimal brain damage ($forgetting OBS$), and self-learning-based neuron importance pruning ($forgetting l2$).
\textbf{(c).} Comparison of the convergence speed of various models. $sgd$ is the valilla model, $forgetting\;layer^{*} $ is the forgetting layer without regularization using the inhibition and lateral inhibition mechanism, and the $forgetting\;layer$ is the complete forgetting layer.}
\label{fig3}
\end{figure*}

The forgetting layer selectively deactivates the insignificant excitatory neurons and consolidates the critical excitatory neurons. To verify its effectiveness in learning the neuron importance, we trained an MLP on Fashion MNIST (784-256-256-128-10 units). We observed the degradation of the model, which corresponds to the fact that deactivating crucial excitatory neurons results in a dramatic decline in the model’s test accuracy and vice versa.
Specifically, we calculate the importance of excitatory neurons (see Methods for details) and rank the corresponding neurons, and then retain a fixed number of neurons according to a given approach, including:
 a positive-order approach, a negative-order approach, and a random approach. The random approach ($random$) is utilized as a control group to reflect the average performance independent of parameter importance. A remarkable difference in performance with $random$ implies that the importance of excitatory neurons is precise. Fig. \ref{fig2} a compares the importance-based manners with the random manners. According to the positive-order approach, the method guarantees the model remains at a high-test accuracy (degradation below 10\%), even removing almost 80\% neurons. It means that the forgetting layer can carefully discard a vast majority of redundant excitatory neurons with minor impairment of learned task-relevant knowledge. In contrast, the model’s performance collapses after removing quite a few neurons (less than 5\%) in the reverse order. It reveals that neural networks with forgetting layers encode information in a limited set of vital excitatory neurons and that destroying them can efficiently collapse the network.

We further analyze the two inactivation modes on various layers, corresponding to neuronal apoptosis in biologically natural amnesia and targeted apoptosis in memory wiping. Fig. \ref{fig2}b reveals that higher-layer neurons are more substantial than those of lower layers, and selective removal of higher-level neurons can be more effective at destroying task-related memories. Relatively, excitatory neurons in the shallow layers are insensitive, suggesting that fewer neurons at the lower layer and shallower networks may benefit the model’s robustness. In addition, the natural forgetting and target forgetting surround a larger area at each layer, further verifying that the forgetting layer can accurately learn the importance of excitatory neurons.

We investigate the distribution of the excitatory neurons’ importance values using the same model as mentioned in Fig. \ref{fig1}e. The plain neural network is taken as a control group to analyse the forgetting layer equipped with a mixture of inhibition and lateral inhibition, as well as  the forgetting layer equipped with an inhibition mechanism. Fig. \ref{fig2}d demonstrates that the forgetting layer substantially facilitates the distribution of excitatory neurons’ importance values close to the bimodal distribution. Moreover, the introduction of the lateral inhibition mechanism further promotes the distribution polarisation. Unlike the near-uniform distribution of the standard model, most values in the distribution with the forgetting layer are concentrated at both ends, i.e., near the zero and the extreme values.

Previous research \cite{krizhevsky2012imagenet,guidotti2018survey,shrikumar2017learning}  holds the view that almost no base neurons dominate the model’s performance in standard convolutional neural networks (CNNs), and groups of convolutional kernels in CNNs perform different functions separately, such as specific neurons responding only to specific features \cite{krizhevsky2012imagenet}. Nevertheless, we demonstrate that CNNs with forgetting layers might possess such base neurons. Experiments on CIFAR-10 \cite{krizhevsky2009learning} with LeNet \cite{lecun1998gradient} suggest that CNNs with a forgetting layer facilitate the formation of base neurons, and removing these neurons leads to dramatic degradation of the model’s performance. Fig. \ref{fig2}c shows the test accuracy of the model after removing the excitatory neurons with $top$-1 and $top$-10 importance. After removing these two neurons, the model fails to recognize almost all categories compared to the original naive standard model.

\subsection*{Forgetting Layers Facilitates Efficient Learning.}
Biologically active forgetting allows for efficient information processing, as exhibited by the adaptability of neural networks to the task and the sparsity of neuronal encoding, preventing information overload \cite{wimber2015retrieval,richards2017persistence}. In particular, for network adaptation, the current widespread use of a uniform network structure ignores data variances, which leads to a mismatch between the network and the data. A straightforward way to address this issue is to manually adjust the number of neurons layer by layer. As a preliminary demonstration, we train a multilayer perceptron network (784-256-256-128-10 units) with configured forgetting layers on Fashion MNIST, whose network complexity is far beyond the data needed (Fig. \ref{fig3}a). The vanilla sgd is the standard model that serves as the control group. Compared with peer algorithms (L1 and L2 regularisation \cite{phaisangittisagul2016analysis}), both of which can reduce the model complexity to fit the data, the use of the forgetting layer can better handle the mismatch between the network size and the data. Quantitatively, the forgetting layer can substantially increase the generalisability of the model, i.e., a nearly 13\% improvement in test accuracy, compared to the vanilla sgd. By comparison, the L1 algorithm has a slight improvement and L2 has a moderate improvement, but they both have a considerable gap with the forgetting layer algorithm.

We next evaluate the ability of our algorithm to store information by probing the pruning parameter size. Intuitively, the fewer parameters the model uses while keeping the test accuracy constant, the more powerful the ability to store information. We trained the VGG network \cite{simonyan2014very} with 9 layers on CIFAR-10. Three parameter pruning strategies are utilised: \textit{(i)} the random, stochastic selection of parameters strategy, which serves as a lower bound on the performance of the algorithm; \textit{(ii)} optimal brain surgery ($OBS$) \cite{hassibi1993second}, one of the mainstream parameter pruning methods, which iteratively selects the parameters with the minimal loss in a pruning-training-pruning manner, and serves as an upper bound on the algorithm; and \textit{(iii)} forgetting layer-based parameter importance sampling ($forgetting\; l2$), which selects the parameter with the lowest importance value, is calculated by the neuron importance as
\begin{equation}\label{eq:2}
    \theta_{i, j}^{l,l-1}=
1-\left(1-\Omega_{i}^{l}\right)\left(1-\Omega_{j}^{l-1}\right)
\end{equation}
where $\Omega_{i}^{l}$ and $\Omega_{j}^{l-1}$ are neurons in adjacent layers. If neither neuron is important, the connection between them, corresponding to the parameter, is even less important, and vice versa. In Fig. \ref{fig3}b, we demonstrate the strong ability of the forgetting layer to store information with a small number of parameters. Specifically, it is capable of maintaining the model's performance with only approximately 20\% of the parameters, close to the level of OBS. It is important to emphasize that our algorithm is online and trained for only one session, while the latter is offline and requires multiple prolonged training-pruning sessions.

In addition, empirical results on CIFAR-10 with VGG suggest that the forgetting layer can accelerate the model convergence. Fig. \ref{fig3}c shows that the model(denoted as $forgetting layer$) with forgetting layers substantially reduces the loss value, and its curve is smoother than the vanilla model's ($sgd$). We further analyse the critical component of the forgetting layer, i.e., the cooperative mechanism of inhibition and lateral inhibition. The results show that the model without this mechanism ($forgetting\;layer^{*} $) has considerably higher loss values and more fluctuating curves than the vanilla model.  We speculate that this is due to the introduction of additional parameters without an effective modulation strategy.

\begin{figure*}[t]
\centering
\includegraphics[width=.9\linewidth]{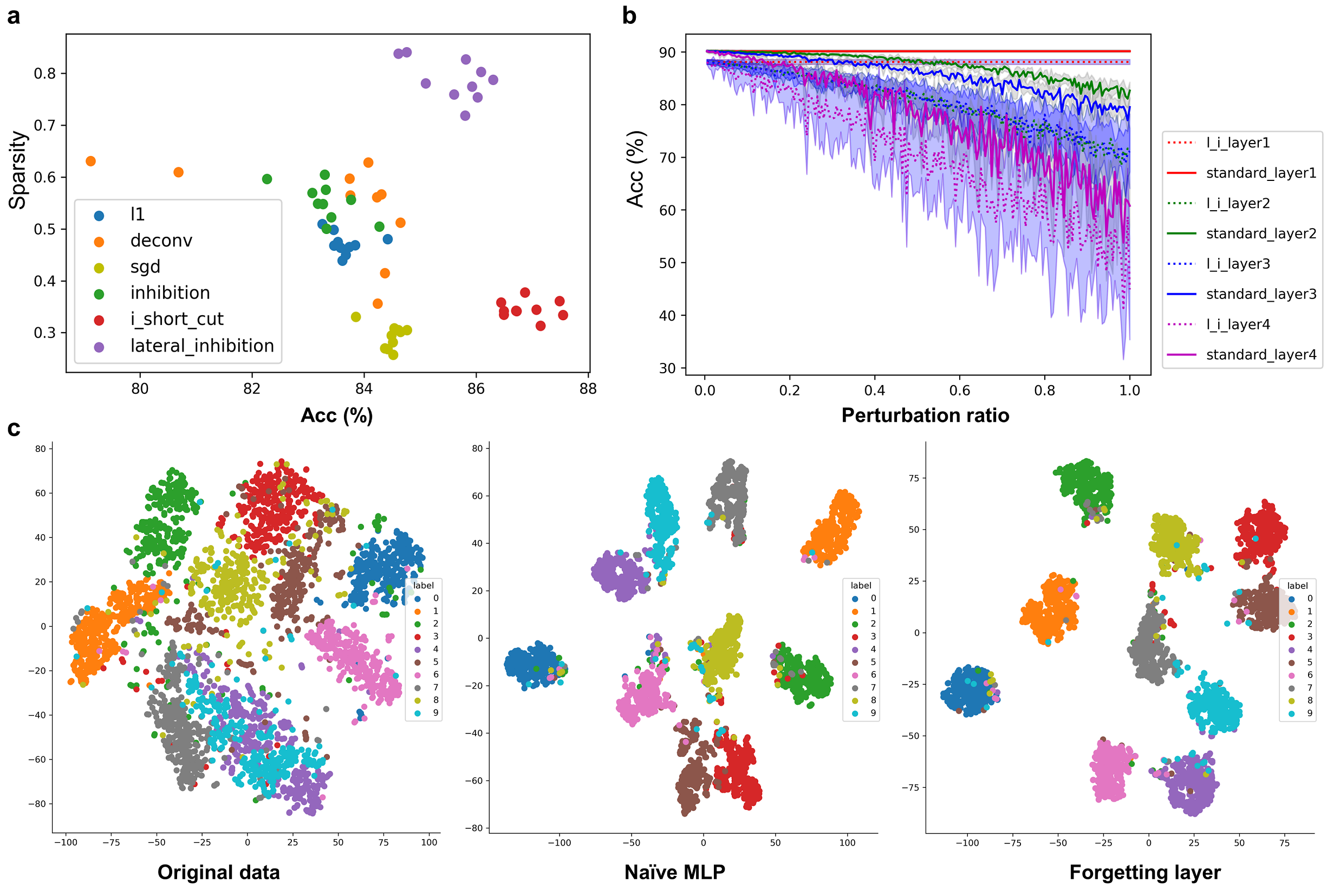}
\caption{Analysis of model generalization and robustness to neuronal perturbations. \textbf{a,}  Scatterplot of model generalization and sparsity. The horizontal axis represents the test accuracy, and the vertical axis represents the sparsity. These results are obtained based on MLP (784-64-64-10) trained ten times repeatedly on Fashion MNIST.  \textbf{b,}  The result of 5 repetitions of random perturbations performed on the neurons of each layer of the model trained on Fashion MNIST. The horizontal axis is the proportion of perturbed neurons, and the vertical axis represents the test accuracy. The solid line indicates the mean value of the test accuracy of the model with forgetting layers and the shaded line indicates the range of the curve’s fluctuation based on the standard deviation calculation. As a comparison, the dashed line represents the mean of the test accuracy of the vanilla neural network. \textbf{c,}  Results of the clustering visualization with $t$-sne. It is based on the features extracted from the vanilla neural network with or without forgetting layers. As a baseline, the original MINST is the result of using images as features.}
\label{fig4}
\end{figure*}

\subsection*{Forgetting Layers Allows for Well-balanced Sparsity, Generalization, and Robustness to the Perturbations.}
Achieving a trade-off between sparsity and generalization is challenging. We show the capability of the forgetting layer in dealing with this problem by training fully connected neural networks on Fashion MNIST. We compared various algorithms, including \textit{(i)} $L1$, one of the classical sparse algorithms; \textit{(ii)} $deconv$ \cite{cogswell2015reducing}, a category of algorithms to reduce overfitting; \textit{(iii)} the vanilla model ($sgd$), a standard neural network; and \textit{(iv)} two critical components in the forgetting layer: the inhibition
mechanism ($inhibition$) and the combination of inhibition and a shortcut strategy ($i\_short\_cut$), as control groups. In Fig. \ref{fig4}a, our algorithm obtains the highest score in sparsity while obtaining suboptimal generalization accuracy. As a comparison, $L1$ and the $deconv$ achieve high sparsity, but both compromise the model’s generalization compared to standard neural networks. Moreover, we analysed the components in the forgetting layer. Although the neural network using only the inhibition mechanism improves the sparsity, it struggles with some loss in generalization accuracy. In contrast, the shortcut combined with inhibition substantially improves the generalization accuracy of the model, but is limited by the low sparsity. Taken together, the model using the complete active forgetting strategy ($lateral\_inhibition$) greatly enhances both generalization accuracy and sparsity.

We further validated the ability of the forgetting layer to improve generalization, especially in terms of features on MNIST to eliminate dataset bias. Specifically, we extracted features from the network’s last layer and clustered them using the $t$-sne algorithm \cite{van2008visualizing}. As baselines, the original input and the features of the vanilla neural network are used. Fig. \ref{fig4}c shows that the forgetting layer enhances the representativeness of the features. Compared with the baselines, the distribution of the features with the forgetting layer is more compact, and the distance between categories is farther. In particular, the distance of distribution between the red points and the brown points is considerably farther than the vanilla model. This indicates a more differentiable feature for forgetting layer learning.

Furthermore, given the above results, we initially investigate whether the forgetting layer can improve robustness, while balancing generalization and sparsity. To achieve this, we randomly perturbed the parameters using the model in Fig. \ref{fig4}a. As a comparison, we also perform a similar operation on the vanilla model. Combining the results of Fig. \ref{fig4}a and Fig. \ref{fig4}b, we show that the forgetting layer can substantially improve the robustness of the model. In particular, the result of separately perturbing each layer supports our conclusion that the sensitivity to perturbation is considerably reduced by utilising the forgetting layer while maintaining high generalization accuracy and sparsity. Moreover, we found that the sensitivity to perturbation varies widely across layers, with shallow layers having lower sensitivity and higher layers having higher sensitivity. The forgetting layer vastly improves the robustness at higher levels, i.e., the difference in generalization accuracy between the forgetting layer and the standard model gradually increases at higher levels.

\begin{figure*}[t]
\centering
\includegraphics[width=.9\linewidth]{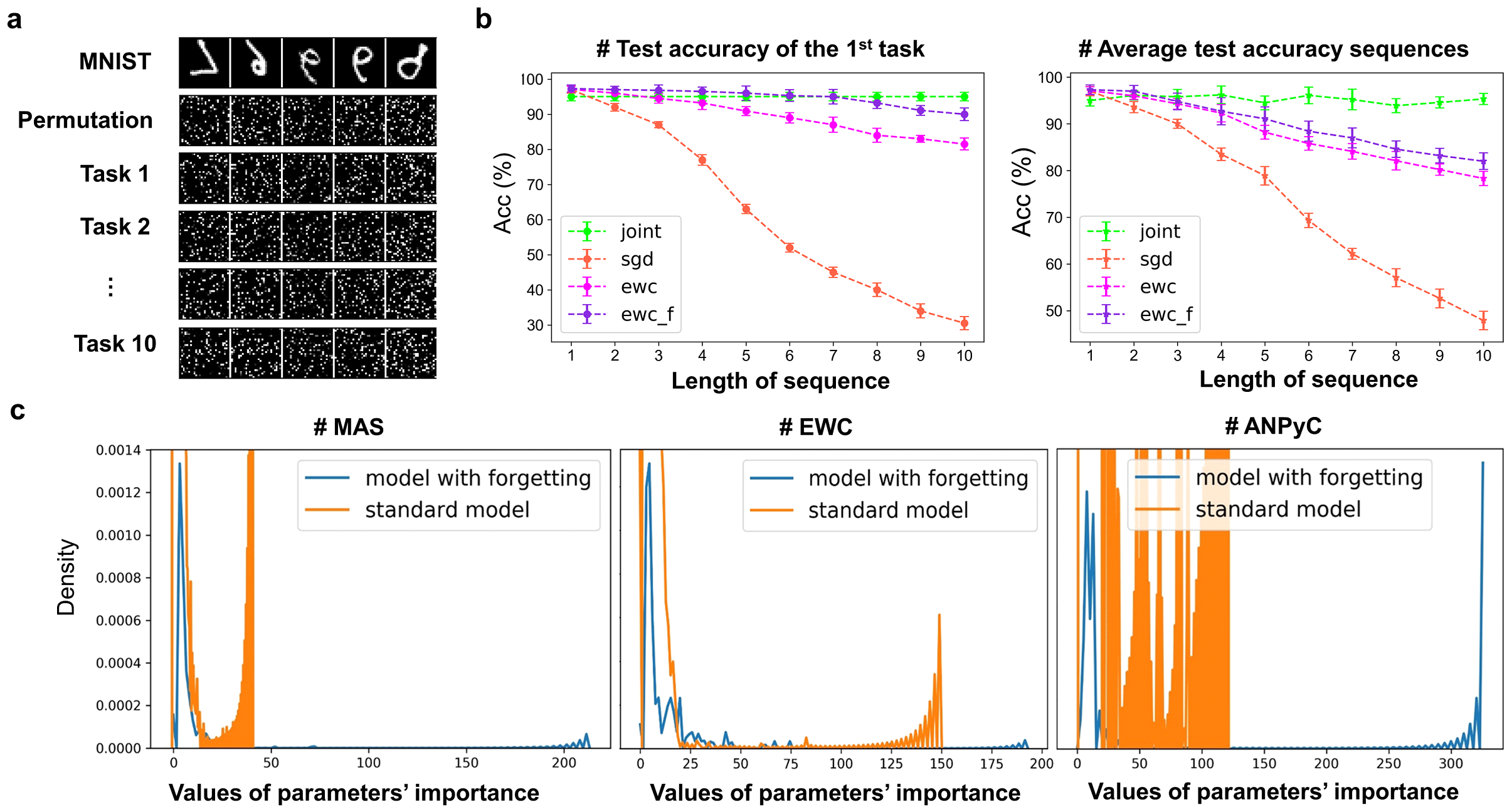}
\caption{Analysis of the forgetting layer for long sequence memory on supervised tasks. \textbf{a,} An example of sequential learning on the PermutedMNIST dataset. \textbf{b,} Results of learning ten tasks sequentially on Permuted MNIST. $ewc$ is EWC based on a vanilla multilayer perceptron model (784-512-256-10 units), $ewc\_f$ is EWC based on the identical multilayer perceptron model equipped with forgetting layers. $joint$ jointly training all tasks at once, free from catastrophic forgetting; $sgd$ sequentially learns tasks one by one in the vanilla model. All tasks share network parameters except for the output layer. \textbf{c,} Comparison of the distribution of parameter-importance values with various methods, on networks with and without forgetting layers. The yellow bars indicate the model using the forgetting layer, and the blue bars indicate the vanilla neural network model.}
\label{fig5}
\end{figure*}

\subsection*{Forgetting Layers Support Long Sequences of Memory.}
The brain learns and remembers continuously by actively forgetting to reduce the interference of future memory \cite{wimber2015retrieval}. Analogously, we expect artificial neural networks to enable continuous learning and memory; however, these networks face a crucial obstacle,  i.e., learning new knowledge and then losing old knowledge, which is known as the catastrophic forgetting issue \cite{mccloskey1989catastrophic}. Studies based on synaptic memory inspire some algorithms to deal with the issue, such as EWC \cite{kirkpatrick2017overcoming} maintaining memory by consolidating parameters related to historical knowledge. Here, we show how forgetting layers can be utilised to improve the ability of such algorithms to overcome catastrophic forgetting and thus achieve long sequence memory. Specifically, we define a long sequence learning scenario in which we sequentially learn subtasks from a given set of tasks (Fig. \ref{fig5}a). The data for these subtasks are Permuted MINIST \cite{srivastava2013compete}, whose samples are obtained by randomly permuting the pixels of the images in MNIST. We train a multilayer perceptron model sequentially on a task set with length $T = 10$, using different strategies, including \textit{(i)} $joint$, which trains all subtasks at once and is an upper bound on model performance; \textit{(ii)} $sgd$, which trains subtasks incrementally using a vanilla model and is a lower bound on model performance; \textit{(iii)} $ewc$, a classical synaptic memory algorithm; and \textit{(iv)} $ewc\_f$, the EWC algorithm that introduces the forgetting layer. Fig. \ref{fig5} b \textit{left} left shows that the forgetting layer can effectively improve the model’s performance in overcoming catastrophic forgetting. During sequential learning, the vanilla model forgets earlier task-related memories, resulting in a sharp degradation of test accuracy. In contrast, the model using $ewc$ and $ewc\_f$ can still maintain the performance of the first task. In particular, the performance of $ewc\_f$ is almost on par with the curve of the joint. It is important to note that the initial test accuracy of $sgd$, $ewc$, and $ewc\_f$ on the first task is higher than that of $joint$ because the former is intended for joint learning of multiple tasks, which is more challenging than learning the first task alone. In addition, we use the average test accuracy to evaluate the ability of the algorithm to memorise long sequences, yielding similar results (Fig. \ref{fig5}b \textit{right}) as Fig. \ref{fig5} b \textit{left}. The results of the two metrics both show that the forgetting layer improves the model’s ability to memorise long sequence tasks. The EWC algorithm, along with the forgetting layer, substantially outperforms the other algorithms. Moreover, the superiority of ($ewc$) increases with the number of tasks in the sequence. The gap between it and the other algorithms gradually increases.

We further explore why the forgetting layer extends the memory duration. Since synaptic memory algorithms centre on measuring the parameter importance, we follow the above experiments by analysing the distribution of parameter importance for algorithms with and without the forgetting layer. In the middle of Fig. \ref{fig5}c \textit{middle}, we first compare the EWC method. We demonstrate that the distribution of parameter importance for the model with the forgetting layer on EWC exhibits polarisation, i.e., the use of the forgetting layer allows the distribution to move towards the ends, with most of the points moving to the zero region, while a smaller fraction of the points is in the larger region. This implies that the forgetting layer allows less model capacity to preserve historical memory while freeing up more capacity for future memory. To exclude the preference of this phenomenon for specific algorithms, we next performed similar statistics on two other synaptic memory algorithms, MAS \cite{aljundi2018memory} and ANPyC \cite{peng2021overcoming} (Fig. \ref{fig5} c \textit{left} and \textit{right}). We demonstrate that it is crucial that the models’ values of parameter importance  using the forgetting layer are more concentrated in the high- and low-value regions. Additionally, there are considerably less points of parameter importance in the intermediate region  than in the model without the forgetting layer, and conversely, points in the zero-value region substantially increase, which dramatically decreases the historical memory-occupied model capacity.

\section*{Discussion}
We first reveal the critical role of active forgetting in learning and memory in this paper.  Then, we propose a model simulating the active forgetting mechanism of the brain based on an artificial neural network called P\&PF. It introduces a novel neuron, called the inhibitory neuron, which acts as a neural switch that selectively regulates the activation state of excitatory neurons. The inhibitory neuron self-regulates the internal state through the IRS, the lateral inhibitory mechanism, and modulates the excitatory neuron through the ERS, the inhibitory mechanism, facilitating the sparsity and representativeness of neural coding. We demonstrate the potential of active forgetting mechanisms on a series of supervised task scenarios, particularly improvements in sparsity, generalization, the robustness of learning, and extended sequence memory.

The forgetting layer introduces selectivity and uncertainty into artificial neural network learning and memorisation. It strengthens highly representational neurons while eliminating redundant neurons, which improve the model’s generalization. Simultaneously, the release of task-irrelevant neurons increases uncertainty, making the learning system more robust to unknown structural and data changes in the future. Moreover, it explicitly expands network capacity, which is the crux of long  persistent memory in sequence of tasks.

Several previous works utilised the concept of forgetting in memory. The work \cite{fusi2005cascade} designed a forgetting module, which implicitly weakened synapses to forget and focuses on the decay of memory with time to model the temporal dynamics, but did not extend to artificial neural networks. And the work \cite{gomez2019learning} filtered information by discarding specific neurons based on activation, but was limited by an effective data selection strategy. In contrast, our algorithm self-learned how to select and discard features by drawing on inhibition and side inhibition mechanisms. This biological mechanism inspired some works. For instance, introducing this mechanism to improve feature representativeness in visual attention and saliency detection \cite{cao2018lateral} or introducing a lateral inhibition mechanism to improve the memory capacity of the model \cite{aljundi2018selfless}, but its parameter importance metric was designed for manually learning, rather than automatically learning like ours.

Although in this paper we focus on an algorithm inspired by biological-based active forgetting mechanisms, we believe that it may help us pursue some important neuroscience issues. In particular, we find that artificial neural networks based on the forgetting layer may exist as a category of neurons that aggregate and encode multilayer features to represent concepts, which are functionally consistent with recently discovered visual nerve cells \cite{han2018logic}. This finding disputes the customary claim that independent neurons encode features in a distributed manner \cite{krizhevsky2012imagenet}. Furthermore, we demonstrate that the forgetting layer performs long-term potentiation (LTP) and long-term depression (LTD)-like functions in brain memory, which are crucial for efficient learning and memory \cite{norimoto2018hippocampal}.

Some problems remain to be discussed and improved in the future.  Specifically,  the global receptive field is utilised  in lateral inhibition. However, a local receptive field combined with a specific prior \cite{turcsany2016local} or an adaptive approach utilising affine transformation \cite{Wei_2017_CVPR} may be beneficial. Moreover, dynamic memory retention based on the forgetting mechanism in long-term memory needs further exploration. To achieve this, it is probably necessary to consider the cycled functional structure of
learning-forgetting. Some directions are deserving of investigation: \textit{(i)} combining attention mechanisms \cite{kim2017joint} to achieve iterative forgetting and recall memory to facilitate learning; \textit{(ii)} introducing active forgetting mechanisms into explicit external memory models \cite{graves2016hybrid} to promote long-term memory consolidation; and \textit{(iii)} drawing on biological neural loop generation mechanisms \cite{Ostapenko_2019_CVPR,mocanu2018scalable} to achieve structural scalability after forgetting.

\section*{Methods}
Inspired by neuroscientific studies, we argue that the crux of biological neural networks to achieve active forgetting lies in \textit{(i)} specific mediators that achieve an indirect regulation of the activation state of excitatory neurons, which is not available in artificial neural networks, and \textit{(ii)} a competition mechanism with a clear division of labor, which acts directly on the mediators to promote a balance between the sparsity of the mediators themselves and the expression of excitatory neurons. Based on these natures, we model the active forgetting mechanism, which introduces inhibitory neurons that act as neural switches to regulate the activation state of excitatory neurons and maintain their sparseness via the IRS while maintaining the representativeness of excitatory neurons via the ERS.

\textbf{Selectivity of forgetting and inhibitory neurons in P\&PF.}
In the "plug-and-play" forgetting layer, we configure specific neural switches called inhibitory neurons to regulate the activation of the corresponding excitatory neurons. It selectively preserves and deactivates activating neurons by a group of weights corresponding to excitatory neurons, which filters out redundant information and preserves critical information. A low weight value means that excitatory neurons are turned off and vice versa. It allows the learning system to encode the task using as few excitatory neurons as possible in artificial neural networks. It allows highly relevant task signals to flow into the next layer, facilitating the learning of representative features while uses a minimum number of excitatory neurons to prevent taking up much of the network capacity.

As shown in Fig. \ref{fig1} a, based on a vanilla artificial neural network, the forgetting layer receives signals from excitatory neurons in the previous layer and modulates the activation of excitatory neurons through inhibitory neurons. It is computed by
\begin{equation}\label{eq:3}
    h(x)^{\prime} = h(x) \odot \sigma
\end{equation}
where $\odot$ is the elementwise dot product operation, $h(x)$ denotes the outputs of the previous layer, and $h(x)^{\prime}$ is the output of the forgetting layer. The $\sigma$ denotes the forgetting layer operation. It regulates the excitatory neurons’ activation flow into the next layer by adjusting inhibitory neurons’ wights. We define the forgetting layer as
\begin{equation}\label{eq:4}
    \sigma=\delta\left(\rho \cdot h(x) \odot W_{f}\right)
\end{equation}

We design the forgetting function $\delta$ that controls the degree of forgetting. To ensure that the algorithm is differentiable, which means that the gradient calculation is available, we introduce a variant of the sigmoid function as the forgetting function, considering that it has the properties of: \textit{(i)} an exponential function whose inverse function is consistent with the Ebbinghaus forgetting curve \cite{ebbinghaus2013memory}, which is consistent with the biological forgetting law; \textit{(ii)} smooth, which guarantees that the algorithm completes end-to-end training by back-propagation; and \textit{(iii)} its range is in  $\left ( 0,1 \right )$, which can prepare to describe the state of the neural switch.

Specifically, on the right side of the equation, $W_{f}$ is the weight of the inhibitory neuron, and $\rho$ is a hyperparameter with a value greater than zero, which controls the degree of polarisation of the sigmoid function. The larger its value is, the steeper the $\delta$ function curve is. Intuitively, $\rho$ acts as a cooling factor, partially affecting the rate of forgetting. For example, increasing the value of $\rho$ will cause more inhibitory neurons to turn on, which means that excitatory neurons turn off and only the most representative excitatory neurons are preserved. On the left side of the equation, $\sigma$ represents the state of the inhibitory neuron, the neural switch, which determines the magnitude of signal flow to the next layer. In the initial state, most neural switches are in a half-on, half-off indeterminate state, and then during the training phase, most task-irrelevant excitatory neurons are turned off through a neural switch-based inhibition mechanism, while only the most critical excitatory neurons survive. Thus, the model learns a $\sigma$ value that reflects the importance of excitatory neurons.

Since the sigmoid function tends to oversaturate with large activation values, this will cause the model to be prone to gradient disappearance on the deep network, making it difficult to converge (Fig. \ref{fig1} b). To alleviate this issue, we introduce the shortcut (Fig. \ref{fig1} a) in the form as
\begin{equation}\label{eq:5}
    h(x)^{\prime}=h(x)+h(x)^{\prime}
\end{equation}

\textbf{The external regulation strategy via inhibition mechanism.}
To ensure that inhibitory neuron weights are learnable, we design an adversarial mechanism.  In addition to making the output of the forgetting layer consistent with the task-relevant truth values, which correspond to the standard loss function, we also keep the inhibitory neurons in the forgetting layer turned on as much as possible, and this corresponds to the loss inspired by the inhibitory mechanism.
The inhibition of neurons realises the neuronal inhibition mechanism. In particular, inhibition neurons inhibit the potential transmission of another group of excitatory neurons through the inhibition mechanism in transmitting signals between neurons. It ensures that minimising inhibitory neurons while maximising task-related information results in a sparse response (Fig. \ref{fig1} d). We give the objective function that inhibits neuron activation. The formula is defined as
\begin{equation}\label{eq:6}
    Loss_{inhibition}=\frac{\sum_{l} \sum_{i}\left(1-\sigma_{i, l}\right) \sigma_{i, l}}{\sum_{l} \sum_{i}\left(1-\sigma_{i, l}\right)}
\end{equation}
where $l$ denotes the $l$-th layer of networks and $i$ represents the location of a neuron in the $l$-th layer.

\textbf{The internal regulation strategy via lateral inhibition mechanism.}
Biological excitatory neurons reduce the activity of neighbouring neurons by a lateral inhibition mechanism. It suppresses the lateral propagation of action potentials from one neuron to its neighbouring neurons. Thus, it drives the contrast of the stimulus signal and then reinforces the encoding of signal. Based on the above mechanism, we built an analogous mechanism for inhibition neurons in the forgetting layer (Fig. \ref{fig1} b) by adding a regularisation in the formula as
\begin{equation}\label{eq:7}
    Loss_{l_{\_}inhibition}=\sum_{l} \sum_{i} \sum_{j}^{i \neq j} h_{i, l} h_{j, l}
\end{equation}
where $l$ denotes the $l$-th layer of neural network, $i$ and $j$ denote neurons' positions in the $l$-th layer.

Furthermore, we consider that in artificial neural networks, the interactions between neurons are related to the neurons’ properties, except to the size of their activation responses. There may be some neurons that are indispensable despite their response values. Therefore, for the lateral inhibition mechanism, we believe it is necessary to consider the importance of neurons and give two hypotheses: \textit{(i)} if one neuron is essential, it should be inhibited less; and \textit{(ii)} if one neuron is essential, it inhibits others more. Based on these two assumptions, we modify the above equation \ref{eq:7} as
\begin{equation}\label{eq:8}
    Loss_{l_{\_}inhibition} = \sum_{l} \sum_{i} \sum_{j}^{i \neq j} h_{i, l} h_{j, l}\left(1-\sigma_{i, l}\right) \sigma_{j, l}
\end{equation}
where $\sigma_{i, l}$ and $\sigma_{j, l}$ are the importance of the two neurons in the $l-th$ layer. We emphasize that it is learned automatically through the excitation-inhibition antagonistic mechanism in the forgetting layer.

\textbf{Forgetting rate and the cooperative constraints of ERS and IRS.}
Combined with the lateral inhibition mechanism, the pair works in concert to actively select and extinguish the insignificant excitatory neurons. The total objective function for configuring the forgetting layer on the standard model is defined as
\begin{equation}\label{eq:9}
    Loss = Loss_{task}+\lambda \cdot Loss_{inhibition}+\beta \cdot Loss_{l_{\_}inhibition}
\end{equation}
where $Loss_{task}$ is the loss function of the task-related objective, and hyperparameters $\lambda$ and $\beta$ control the weights of inhibition and lateral inhibition, as well as the rate of forgetting, respectively.

\textbf{Implementation.} The implementation of the training and testing models equipped with active forgetting is based on the framework, TensorFlow \cite{abadi2016tensorflow}.


\section*{Acknowledgements}

This work was supported in part by the National Natural Science Foundation of China under Grant 41871364, Grant 41571397, Grant 42071427, and Grant 41771458 and by using Computing Resources at the High Performance Computing Platform of Central South University.

\section*{Author contributions statement}



\section*{Ethics declarations}

Competing interests:

The authors declare no competing interests.

\section*{Additional information}
The source code and data will be available at https://github.com/GeoX-Lab/P-PF.

\end{document}